\documentclass[10pt,twocolumn,letterpaper]{article}

\usepackage{times}
\usepackage{epsfig}
\usepackage{graphicx}
\usepackage{amsmath}
\usepackage{amssymb}
\usepackage{enumerate}
\usepackage{paralist}
\usepackage[font=small,labelfont=bf,tableposition=top]{caption}

\usepackage{tikz}

\usepackage{accents}
\newlength{\dhatheight}
\newcommand{\doublehat}[1]{%
    \settoheight{\dhatheight}{\ensuremath{\hat{#1}}}%
    \addtolength{\dhatheight}{-0.25ex}%
    \hat{\vphantom{\rule{1pt}{\dhatheight}}%
    \smash{\hat{#1}}}}

\def\x{\mathbf{x}}
\def\y{\mathbf{y}}
\def\z{\mathbf{z}}

\def\hx{{\hat{\mathbf{x}}}}
\def\hy{{\hat{\mathbf{y}}}}
\def\hhx{{\doublehat{\mathbf{x}}}}
\def\hhy{{\doublehat{\mathbf{y}}}}
\def\hB{{\hat{B}}}
\def\hO{{\hat{O}}}
\def\hhB{{\doublehat{B}}}

\def\m{\mathbf{m}}

\def\mask{\mathrm{Mask}}
\def\gs{G_{\mathrm{seg}}}
\def\gi{G_{\mathrm{inp}}}
\def\ge{G_{\mathrm{enh}}}
\def\discrb{D_{\mathrm{bg}}}
\def\discro{D_{\mathrm{obj}}}
\def\l{\mathcal{L}}
\def\ldb{\l^{\mathrm{GAN}}_{\mathrm{bg}}}
\def\lrb{\l^{\mathrm{rec}}_{\mathrm{bg}}}
\def\ldo{\l^{\mathrm{GAN}}_{\mathrm{obj}}}
\def\lro{\l^{\mathrm{rec}}_{\mathrm{obj}}}
\def\lio{\l^{\mathrm{id}}_{\mathrm{obj}}}
\def\lib{\l^{\mathrm{id}}_{\mathrm{bg}}}
\def\ldiscrb{\l^{\mathrm{disc}}_{\mathrm{bg}}}
\def\ldiscro{\l^{\mathrm{disc}}_{\mathrm{obj}}}
\usepackage{tabularx}

\usepackage[pagebackref=true,breaklinks=true,letterpaper=true,colorlinks,bookmarks=false]{hyperref}

\begin{document}

%%%%%%%%% TITLE
\title{SEIGAN: Towards Compositional Image Generation by \\ Simultaneously Learning to Segment, Enhance, and Inpaint}

\author{Pavel Ostyakov$^1$ \and Roman Suvorov$^1$
	\and Elizaveta Logacheva$^1$ \and Oleg Khomenko$^1$
	\and Sergey I. Nikolenko$^{1,2,3}$\\
	\\
	$^1$Samsung AI Center, Moscow, Russia\\
	\texttt{\small \{p.ostyakov, r.suvorov, e.logacheva, o.khomenko\}@samsung.com} \\
	$^2$Steklov Mathematical Institute at St. Petersburg\\
	$^3$Neuromation OU, Tallinn, Estonia\\
	\texttt{\small sergey@logic.pdmi.ras.ru}\\	
	\\
}

\twocolumn[{%
\renewcommand\twocolumn[1][]{#1}%
\vspace{-3em}
\maketitle

\vspace{-3em}

\begin{center}
    \includegraphics[width=\linewidth]{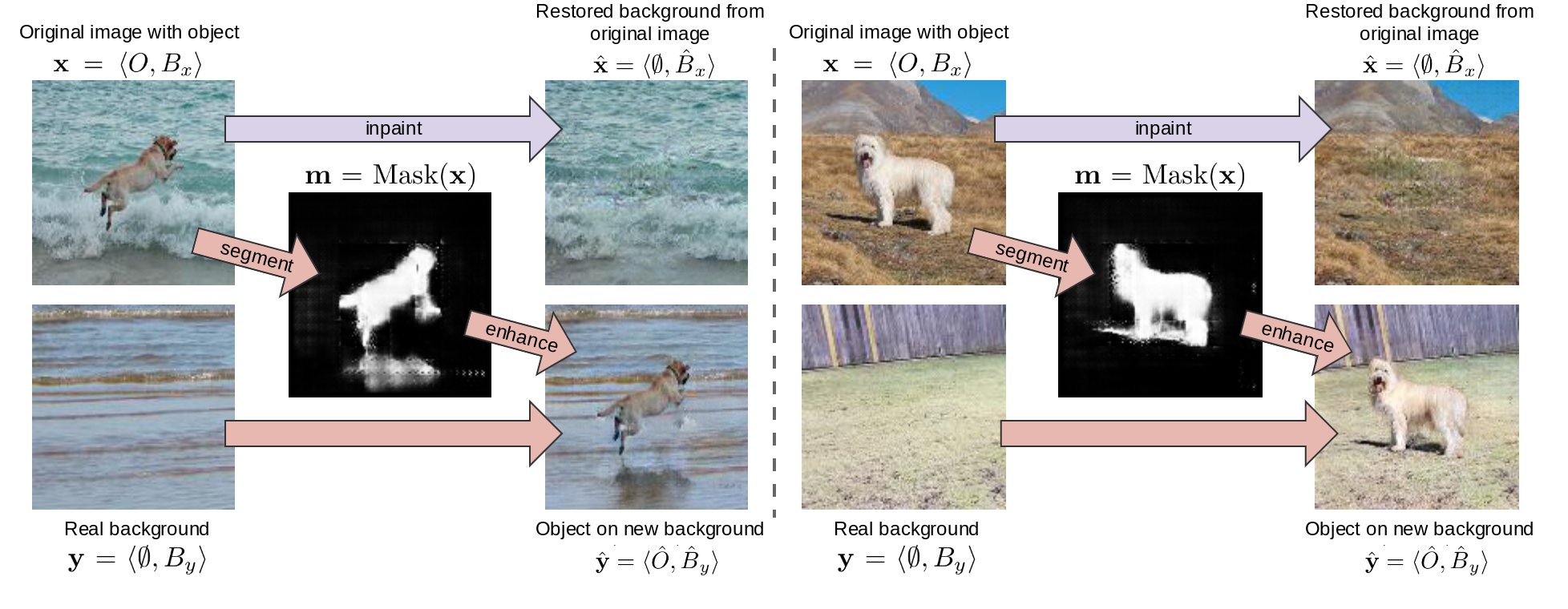}
    % \vspace{-0.5em}
    \captionof{figure}{
    Illustration of the problem and our approach: given two images, $\x$ with object and $\y$ with background, generate two images: $\hx$ with background from the first image restored after cutting out the object and $\hy$ with the object pasted onto the second background. Images on this figure are real examples produced by our algorithm.
    }\label{fig:problem}
\end{center}

}]

%%%%%%%%% ABSTRACT

\begin{abstract}
We present a novel approach to image manipulation and understanding by simultaneously learning to segment object masks, paste objects to another background image, and remove them from original images. For this purpose, we develop a novel generative model for compositional image generation, SEIGAN (Segment-Enhance-Inpaint Generative Adversarial Network), which learns these three operations together in an adversarial architecture with additional cycle consistency losses.
% to improved quality of segmentation masks and better perceptual quality of generated images. 
To train, SEIGAN needs only bounding box supervision and does not require pairing or ground truth masks. SEIGAN produces better generated images (evaluated by human assessors) than other approaches and produces high-quality segmentation masks,
% against ground truth masks on the COCO dataset, 
improving over other adversarially trained approaches and getting closer to the results of fully supervised training.
% (0.762 vs. IoU of 0.83 achieved by ResNet trained with full supervision and 0.723 by simple adversarially trained segmentation).
\end{abstract}

% \begin{figure*}[t]\centering
%     \includegraphics[width=\linewidth]{images/teaser.png}
%     % \vspace{-0.5em}
%     \caption{
%     Illustration of the problem and our approach: given two images, $\x$ with object an and $\y$ with background, generate two images: $\hx$ with background from the first image restored after cutting out the object and $\hy$ with the object pasted onto the second background. Images on this figure are real examples produced by our algorithm.
%     }\label{fig:problem}
% \end{figure*}

%%%%%%%%% BODY TEXT
\section{Introduction}

Image generation, an important problem in itself, is also commonly viewed as a stepping stone for machine learning models with deeper understanding. Current state of the art approaches are usually based on deep generative models: generative adversarial networks (GANs) and variational autoencoders (VAE). Both approaches still have a lot of drawbacks, including image quality and consistency, training stability, variability, and level of control over the generation process. However, GANs and VAE solve these problems better and better, and over the last few years have gained huge popularity as a powerful and flexible (but difficult to use) tool for image generation, style transfer, domain adaptation, inpainting (image restoration), weakly supervised object segmentation, and other similar applications.

An alternative approach might be \emph{compositional image generation}: construct a new image from parts of other images instead of trying to make a new scene from scratch in full detail. In this work, we make a step towards compositional image generation, learning to construct a new image given two source images, one with an object and the other with a background. This process consists of three basic operations:
\begin{inparaenum}[(1)]
\item \textit{cut}, extracting an object from image;
\item \textit{paste and enhance}, making the pasted object appear natural in the new context;
\item \textit{inpaint}, restoring the background after cutting out an object;
\end{inparaenum}
see Fig.~\ref{fig:problem} for an illustration.

To perform these operations, we propose a novel neural network architecture, SEIGAN (Segment-Enhance-Inpaint GAN), which allows to train neural networks for all three operations simultaneously, in an end-to-end fashion. The training process is weakly supervised,
% Currently, it requires an object detector for data preparation, but we are working on overcoming this limitation. 
and we show that learning all these operations together allows to improve the perceptual quality of final images as well as the accuracy of segmentation masks (a side product of this problem, which may also become a direct application for compositional image generation models~\cite{remez2018learning}).

The paper is organized as follows: in Section~\ref{sec:related} we review recent important work from relevant areas, in Section~\ref{sec:methods} we define the SEIGAN model formally, Section~\ref{sec:eval} presents an experimental evaluation and discusses the performance of our model, and Section~\ref{sec:concl} concludes the paper.

%------------------------------------------------------------------------
\section{Related Work}\label{sec:related}

\textbf{Unsupervised and weakly supervised object segmentation.} 
Traditional approaches to unsupervised image segmentation have used superpixel clustering~\cite{10.1117/1.JEI.26.6.061610}, but recently this problem has been addressed with deep learning. In~\cite{ji2018invariant}, the network learns to maximize information between two cluster-id vectors obtained by a fully convolutional network from nearby patches of the same image. A similar technique, but constrained with the reconstruction loss, has been proposed in~\cite{xia2017w}, where the \emph{W-Net} architecture (an autoencoder with U-Net-like encoder and decoder) learns to cluster pixels at the inner layer and then reconstruct an image from pixel clusters. Their segmentation result is unaware of object classes. 
% The work~\cite{wu2018annotation} proposes an annotation-free framework to learn a segmentation network for homogeneous objects; it uses an adaptive synthetic data generation process to create a training dataset. 
The works~\cite{Zhang2017SupervisionBF,jing2018deep} use weak salience maps from several handcrafted detectors as supervision for a DNN-salient object detector.

Perhaps the nearest to our present work is~\cite{remez2018learning}, where a GAN-based approach~\cite{zhu2017unpaired} is used to generate object segmentation masks from bounding boxes by learning to cut and paste objects. The training pipeline consists of taking two crops of the same image: one with object and one without any object, as detected with Faster R-CNN~\cite{ren2015faster}. Then a GAN architecture is trained to produce a segmentation mask $\m$ so that these two crops merged with $\m$ result into a plausible image. The loss function in~\cite{remez2018learning} is a combination of adversarial loss, existence loss (that verifies that an object is present on an image) and cut loss (that verifies that no part of the object is left after it has been cut); 
% experiments cover some classes from Cityscapes~\cite{Cordts2016Cityscapes} and all classes from COCO~\cite{lin2014microsoft} datasets. 
the authors report that their approach achieves higher mean intersection-over-union (mIoU) values than the classic \emph{GrabCut} algorithm~\cite{rother2004grabcut} and recent \emph{Simple-Does-It} approach~\cite{khoreva2017simple}. However, this approach requires a pretrained Faster R-CNN and a special policy for foreground and background patch selection, experiences difficulties with some object classes, and works well only for small resolution images ($28 \times 28$).
 
\textbf{Visual grounding.} Methods for visual grounding aim to perform unsupervised or weakly supervised matching of free-form text queries and regions of images. Usually supervision takes form of $(\mathrm{Image}, \mathrm{Caption})$ pairs. The model performance is usually measured as intersection-over-union (IoU) against ground truth labels, and the most popular datasets for evaluation include Visual Genome~\cite{krishnavisualgenome}, Flickr30k~\cite{plummer2015flickr30k}, Refer-It-Game~\cite{kazemzadeh2014referitgame}, and COCO~\cite{lin2014microsoft}. A general approach to grounding is to predict if a given caption and image correspond to each other, obtaining negative samples by assigning random captions. Text-image attention is the core feature of most models for visual grounding~\cite{yu2018mattnet}. Obviously, using more fine-grained supervision (e.g. region-level annotations instead of image-level) leads to better performance~\cite{zhang2017discriminative}.

\textbf{Trimap generation.} Trimap generation is the problem of segmenting an image into three classes: foreground, background, and unknown (transparent foreground). Most algorithms require human intervention, but recently superpixel- and clustering-based approaches have been proposed for automated trimap generation~\cite{gupta2016automatic}. However, their approach requires multiple optimization steps for every image. Deep neural networks have been used to produce an alpha matting mask given an image and a trimap~\cite{xu2017deep}, and for video matting and background substitution in video~\cite{huang2017practical}, where results of per-frame superpixel segmentation are joined in a conditional random field of Gaussian mixture models that learns to separate foreground and background frame-by-frame.

\textbf{Generative adversarial networks.} In recent years, GANs~\cite{goodfellow2014generative} are probably the most frequently used approach to train a generative model. Though powerful, they are prone to instability in training and inconsistent performance on higher resolution images. The recently proposed CycleGAN~\cite{zhu2017unpaired} trains two GANs together to establish a bidirectional mapping between two domains. It achieves much greater stability and consistency by using the intuition that after a full cycle, going from one domain to another and back, the image should remain unchanged; this requires the underlying operation to be invertible in some sense. Many modifications and applications of CycleGAN have already been proposed, including semantic image manipulation~\cite{wang2017high}, domain adaptation~\cite{bousmalis2018using}, unsupervised image-to-image translation~\cite{liu2017unsupervised}, multi-domain translation~\cite{choi2017stargan}, and many others. Another problem is that such a mapping between domains may be ambiguous; BicycleGAN~\cite{zhu2017toward} and augmented CycleGAN~\cite{almahairi2018augmented} address this problem by requiring that the mapping must preserve latent representations.

In this work, we build upon the ideas of cut-and-paste~\cite{remez2018learning} and CycleGAN~\cite{goodfellow2014generative} and propose a novel architecture for background swapping, a variation of the cut-and-paste problem that can be thought of as invertible. Our model achieves better results on unsupervised object segmentation, inpainting, and image blending.

\section{Methods}\label{sec:methods}

% \begin{figure}\centering
%     \includegraphics[width=\linewidth]{images/seigan-1.png}
    
%   \caption{Problem setting.}
% \label{fig:problem}
% \end{figure}

\subsection{Problem Setting and General Pipeline}

Throughout the paper, we denote images as object-background tuples, e.g. $\x=\langle O, B_x\rangle$ means that image $\x$ contains object $O$ and background $B_x$, and $\y=\langle \emptyset, B_y \rangle$ means that image $\y$ contains background $B_y$ and no objects.

The main problem that we address in this work can be formulated as follows. Given a dataset of background images $Y = \left\{ \langle \emptyset, B_y \rangle\right\}_{\y\in Y}$ and a dataset of objects on different backgrounds $X= \left\{ \langle O_x, B_x \rangle\right\}_{\x\in X}$ (unpaired, i.e., with no mapping between $X$ and $Y$), train a model to take an object from an image $\x\in X$ and paste it onto a new background defined by an image $\y\in Y$, while at the same time deleting it from the original background. In other words, the problem is to transform a pair of images $ \x=\langle O, B_x \rangle $ and $ \y=\langle \emptyset, B_y \rangle$ into a new pair $ \hx= \langle \emptyset, \hB_x \rangle$ and $\hy= \langle \hO, \hB_y \rangle$, where $\hB_x\approx B_x$, $\hB_y\approx B_y$, and $\hO\approx O$, but the object and both backgrounds are changed so that the new images look natural. An optional bonus would be to also output the segmentation mask $\m$ for object $O$ in $\x$. We keep the same notation throughout the paper, tying the letters $\x$ and $\y$ to the backgrounds; e.g., $\x=\langle O, B_x \rangle$ is the original image, and $\hat\x$ is the result of removing $O$ from $\x$ and inpainting; Fig.~\ref{fig:problem} illustrates the problem and our basic notation.

% {\bf WOULD BE GREAT TO HAVE A PICTURE ILLUSTRATING THE PROBLEM SETTING}
% \section{Method}
% \subsection{General Pipeline}
The problem can be decomposed into three subtasks:
\begin{itemize}
    \item \emph{segmentation}: segment the object $O$ from an original image $\x= \langle O, B_x \rangle $ by predicting the segmentation mask $\m=\mask(\x)$; given the mask, we can make a coarse blend that simply cuts off the segmented object from $\x$ and pastes it onto $\y$: $\z=\m\odot\x + (1-\m)\odot \y$, where $\odot$ denotes componentwise multiplication;
    \item \emph{enhancement}: given  original images $\x$ and $\y$ (in our case, the corresponding part of the network accepts the coarse image $\z$ as input) and segmentation mask $\m$, construct an enhanced version $\hy=\langle \hO, \hB_y \rangle$;
    \item \emph{inpainting}: given 
    % a segmentation mask $\m$ and 
    an image $(1-\m)\odot \x$ obtained by zeroing out pixels of $\x$ according to $\m$, restore the background-only image $\hx=\langle\emptyset, \hB_x\rangle$.
\end{itemize}

Each of these problems has been, in some form or another, addressed in literature (see Section~\ref{sec:related}), and for each of these tasks we can construct a separate neural network that accepts an image or a pair of images and outputs new image or images of the same dimensions.
 % Segmentation and enhancement networks constitute a bigger module, which we call "Swap network".
However, our main hypothesis that we explore in this work is that in the absence of large paired and labeled datasets (which is the normal state of affairs in most applications), it is highly beneficial to train all these neural networks together.

\begin{figure*}
    \begin{center}
        \includegraphics[width=\linewidth]{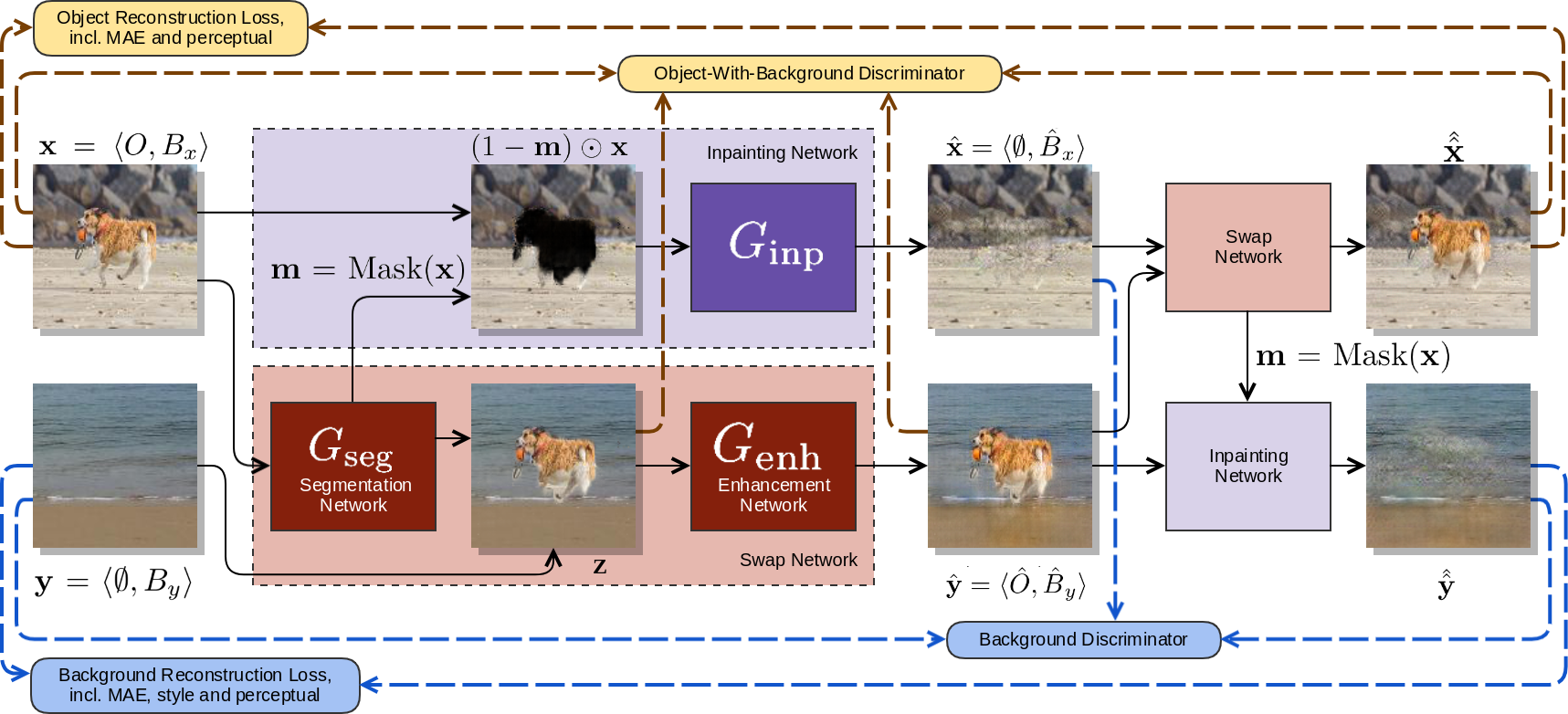}
    \end{center}
    
   \caption{A high-level overview of the SEIGAN pipeline for joint segmentation and inpainting: the swap and inpainting networks are applied twice to complete the cycle. Solid rectangles denote neural networks; rounded rectangles, loss functions; solid lines show data flows, and dashed lines indicate the flow of values to loss functions.}
\label{fig:architectures}
% \end{figure*}

% % \begin{figure*}[t]
%     \begin{center}
%         \includegraphics[width=0.8\linewidth]{images/swap-network-narrow.png}
%     \end{center}
    
%   \caption{Architecture of the swap network that cuts the object from one image and pastes it onto another. 
%   % The enhancement network in its turn can be recursively expanded with two steps of segmentation and more fine-grained enhancement.
%   }
% \label{fig:swap-network}
\end{figure*}

Thus, we present our SEIGAN (Segment-Enhance-Inpaint) architecture that combines all three components in a novel and previously unexplored way. We outline the general flow of our architecture on Figure~\ref{fig:architectures}; the ``swap network'' module there combines segmentation and enhancement. Since cut-and-paste is a partially reversible operation, we organize the training procedure in a way similar to CycleGAN~\cite{zhu2017unpaired}: swap and inpainting networks are applied twice in order to use the idempotency property for the loss functions. We denote by $\hx$ and $\hy$ the results of the first application, and by $\hhx$ and $\hhy$ the results of the second application, moving the object back from $\hy$ to $\hx$ (see Fig.~\ref{fig:architectures}).

The architecture combines five different neural networks, three used as generators and two as discriminators:
\begin{itemize}
    \item $\gs$ is solving the segmentation task: given an image $\x$, it predicts $\mask(\x)$, the segmentation mask of the object on the image;
    \item $\gi$ is solving the inpainting problem: given $(1-\m)\odot \x$, predict $\hx=\langle\emptyset, \hB_x\rangle$;
    \item $\ge$ does enhancement: given 
    % $\x$, $\y$, and 
    $\z=\m\odot\x + (1-\m)\odot \y$, predict $\hy=\langle \hO, \hB_y \rangle$;
    \item $\discrb$ is the background discriminator that attempts to distinguish between real and fake (inpainted) background-only images; its output $\discrb(\x)$ should be close to $1$ if $\x$ is real and close to $0$ if $\x$ is fake;
    \item $\discro$ is the object discriminator that does the same for object-on-background images; $\discro(\x)$ should be close to $1$ if $\x$ is real and close to $0$ if $\x$ is fake.
\end{itemize}
Generators $\gs$ and $\ge$ constitute the so-called ``swap network'' (red shaded rectangle on Fig.~\ref{fig:architectures}).
% depicted as a single unit on Fig.~\ref{fig:high-level} and explained in detail on Fig.~\ref{fig:swap-network}.
%
Compared to~\cite{remez2018learning}, the training procedure in SEIGAN has proven to be more stable and able to work in higher resolutions, and the resulting segmentation masks are better. Furthermore, our architecture allows to address more tasks (inpainting and blending) simultaneously rather than only predicting segmentation masks. As usual in GAN design, the secret sauce of the architecture lies in a good combination of different loss functions. In SEIGAN, we use a combination of adversarial, reconstruction, and regularization losses; over the following sections, we cover each part in detail.

\begin{figure*}[t]
    \begin{tabular}{c|c}
        \includegraphics[width=0.48\linewidth]{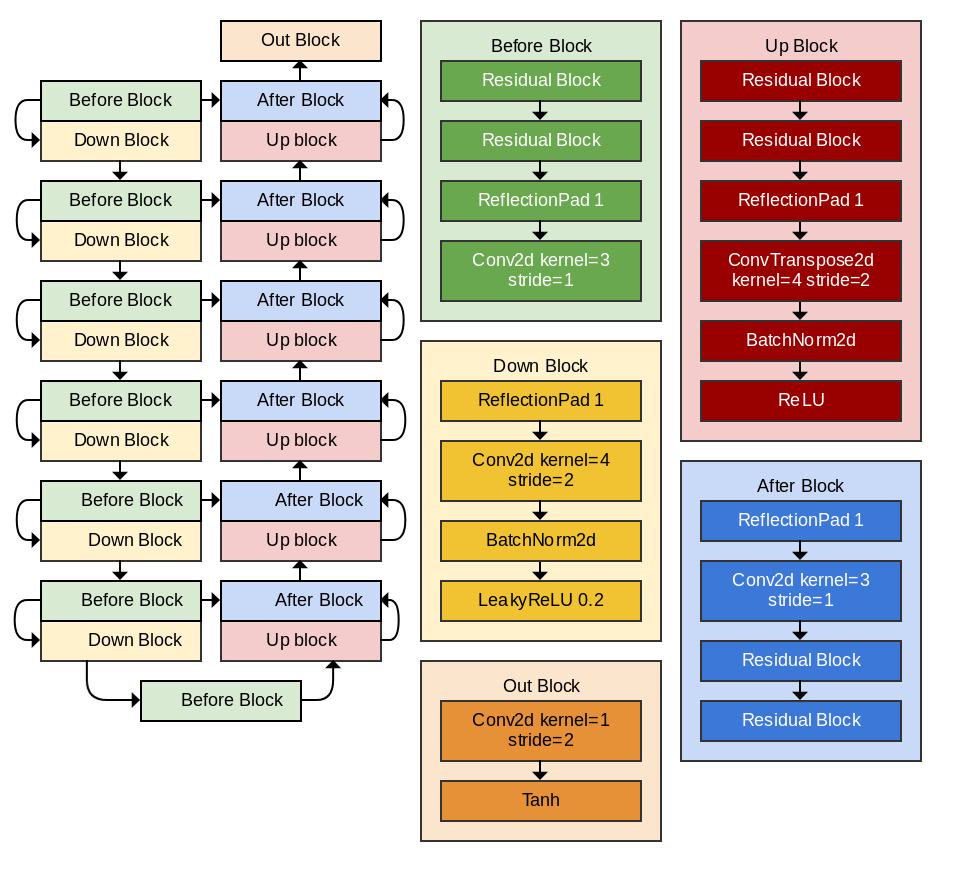} &
        \includegraphics[width=0.48\linewidth]{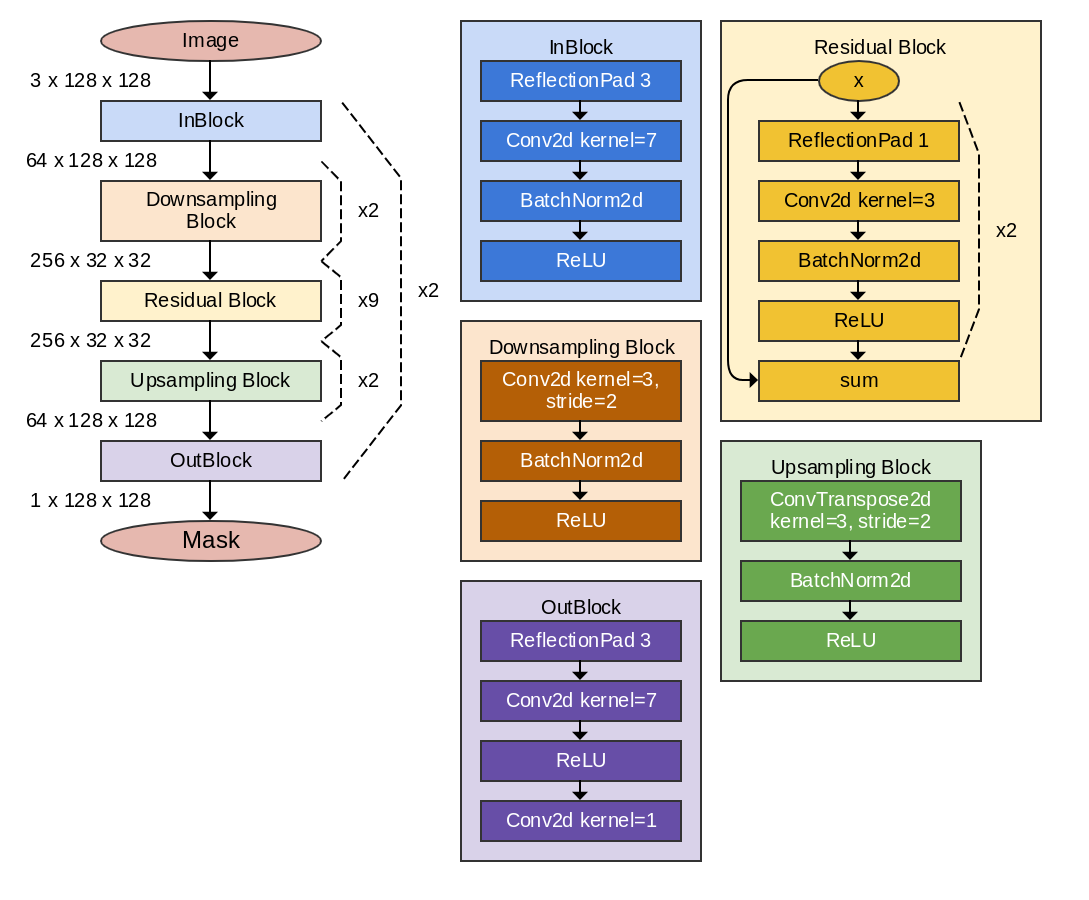}
    \end{tabular}

   \caption{Architectures U-Net (left) and ResNet (right).
   U-Net performed best for inpainting and refinement networks, while ResNet powered our most successful segmentation network.
   }
\label{fig:arch}

\end{figure*}

\subsection{The Inpainting Network $\gi$}

The inpainting network $\gi$ aims to produce a plausible background $\hB_x$ given a random noise sampled from standard normal distribution and a source image $(1-\m)\odot \x$, which represents the original image $\x$ with the object subtracted according to segmentation mask $\m$ obtained by applying the segmentation network, $\m=\gs(\x)$; in practice, we zero out the pixels of $\m\odot\x$. Parameters of inpainting network are optimized during the end-to-end training according to the following loss functions (shown by rounded rectangles on Fig.~\ref{fig:architectures}).

% \begin{itemize}
The \emph{adversarial background loss} aims to improve the plausibility of the resulting image with a dedicated discriminator $\discrb$ with the ImageGAN architecture as proposed in~\cite{chu2017cyclegan} except for the number of layers; our experiments have shown that a deeper discriminator works better. As the loss function $\discrb$ uses the MSE adversarial loss suggested in Least Squares GAN (LSGAN)~\cite{mao2016least}, as in practice it is by far more stable than other types of GAN loss functions. Therefore, for generator networks we used

\noindent
\begin{align*}
{\l_{\mathrm{inp}}^{\mathrm{GAN}}} &= (1 - \discrb(\hx))^2,\quad {\l_{\mathrm{inp2}}^{\mathrm{GAN}}} = (1 - \discrb(\hhy))^2,\\
  \ldb &= \lambda_1{\l_{\mathrm{inp}}^{\mathrm{GAN}}} +
  \lambda_2{\l_{\mathrm{inp2}}^{\mathrm{GAN}}},
\end{align*}
% 
% \noindent
where $\hx=\langle \emptyset, \hB_x \rangle$ is the background image resulting from $\x$ after the first swap, $\hhy=\langle \emptyset, \hhB_y \rangle$ is the background image resulting from $\hy$ after the second swap, and $\lambda_i$ here and below are constants to be determined empirically.

The \emph{background reconstruction loss} $\lrb$ aims to preserve information about the original background $B_x$. It is implemented using \emph{texture loss}~\cite{DBLP:journals/corr/XianSLFYH17}, the mean average difference between Gram matrices of feature maps after the first 5 layers of VGG-19 network, and \emph{perceptual loss}, the mean average difference between feature maps after the last 5 layers of VGG-19 network, and pixel-level MAE loss:
\begin{align*}
  {\l_{\mathrm{bg}}^{\mathrm{texture}}} &= \left| \mathrm{Gram}(\mathrm{VGG_1}(\y)) - \mathrm{Gram}(\mathrm{VGG_1}(\hhy)) \right|,\\
  {\l_{\mathrm{bg}}^{\mathrm{perc}}} &= \left| \mathrm{VGG_2}(\y) - \mathrm{VGG_2}(\hhy) \right|,\quad
  {\l_{\mathrm{bg}}^{\mathrm{MAE}}} = \left| \y - \hhy \right|,\\
  \lrb &= \lambda_3{\l_{\mathrm{bg}}^{\mathrm{texture}}} +
  \lambda_4{\l_{\mathrm{bg}}^{\mathrm{perc}}} + 
  \lambda_5{\l_{\mathrm{bg}}^{\mathrm{MAE}}},
\end{align*}
% $$$$
% $$$$
% $$,$$
% $$\lrb = \lambda_{\mathrm{rec_{texture}}} {\l^{\mathrm{rec_{texture}}}_{\mathrm{bg}}} + \lambda_{\mathrm{rec_{percept}}} {\l^{\mathrm{rec_{percept}}}_{\mathrm{bg}}} + \lambda_{\mathrm{rec_{mae}}} {\l^{\mathrm{rec_{mae}}}_{\mathrm{bg}}},$$

\noindent
where $\mathrm{VGG_1}(\y)$ denotes the matrix of features after the first 5 convolutional layers in a pretrained VGG-19 network, $\mathrm{VGG_2}(\y)$, after the last 5 convolutional layers, and $\mathrm{Gram}(A)_{ij} = \sum_{k}A_{ik}A_{jk}$ is the Gram matrix.

% Our choice of loss functions is motivated by the fact that 
Note that there are plenty of possible plausible reconstructions of the background, so the loss functions must allow for a certain degree of freedom that mean absolute error or mean squared error would not permit but which texture and perceptual losses do. In our experiments, optimizing only MAE or MSE has usually led to the generated image being filled with median or mean pixel values, with no objects or texture. Note that the background reconstruction loss is applied only to $\y$ because we do not have the ground truth background for $\x$ (see Fig.~\ref{fig:architectures}). 

Note that before feeding the image to the inpainting network $\gi$, we subtract a part of image according to the mask $\m$ in a differentiable way, without any thresholding applied to $\m$. Thus, gradients can propagate back through $\m$ to the segmentation network $\gs$. Joint training of inpainting and segmentation has a regularization effect. First, the inpainting network $\gi$ wants the mask to be as accurate as possible: if it is too small then $\gi$ will have to erase the remaining parts of the objects, which is a harder problem, and if it is too large then $\gi$ will have more empty area to inpaint. Second, $\gi$ wants the segmentation mask $\m$ to be high-contrast (with values close to $0$ and $1$) even without thresholding: if much of $\m$ is low-contrast (close to $0.5$) then $\gi$ will have to learn to remove the ``ghost'' of the object (again, harder than just inpainting on empty space), and it will most probably be much easier for the discriminator $\discrb$ to tell that the resulting picture is fake.
For $\gi$, we use a neural network consisting of two U-Net networks connected sequentially (see Fig.~\ref{fig:arch}).

\subsection{The Swap Network}

The swap network aims to generate a new image $\hy=\langle \hO, \hB_y \rangle$ given two images, $\x=\langle O, B_x\rangle$ with an object $O$ and $\y=\langle \emptyset, B_y \rangle$ with a different background $ B_y$. The swap network consists of two major steps: segmentation $\gs$ and enhancement $\ge$ (red shaded rectangle on Fig.~\ref{fig:architectures}).

The \emph{segmentation network} $\gs$ produces a soft segmentation mask $\m=\gs(\x)$, that can be used to extract the object $O$ from $\x$ and paste it on $B_y$ to produce a ``coarse'' version of the target image $\z=\m\odot\x + (1-\m)\odot \y$. But $\z$ is not the end result: it lacks anti-aliasing, color/lightning correction, and other improvements. Note that pasting an object in a perfectly natural way might require a very involved understanding of the background; e.g., when pasting a dog onto a grass field we should probably hide some part of its paws behind grass as they would be hidden in reality.

\begin{figure*}
\renewcommand{\arraystretch}{0.6}
\setlength{\tabcolsep}{1.0pt}
\begin{center}
\begin{tabular}{ccccccc}
    Method & $\x$ & $\y$ & $\z$ & $\hy$ & $\hx$ & $\m$  \\
\hline
    \rotatebox{90}{$\qquad$SEIGAN}
        & \includegraphics[width=0.15\linewidth]{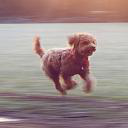}
        & \includegraphics[width=0.15\linewidth]{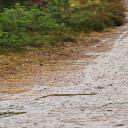}
        & \includegraphics[width=0.15\linewidth]{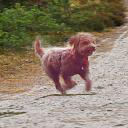}
        & \includegraphics[width=0.15\linewidth]{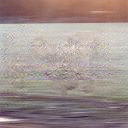}
        & \includegraphics[width=0.15\linewidth]{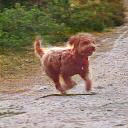}
        & \includegraphics[width=0.15\linewidth]{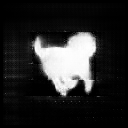}
        \\
    \rotatebox{90}{$\qquad$SEIGAN}
        & \includegraphics[width=0.15\linewidth]{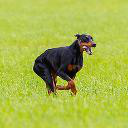}
        & \includegraphics[width=0.15\linewidth]{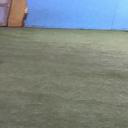}
        & \includegraphics[width=0.15\linewidth]{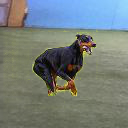}
        & \includegraphics[width=0.15\linewidth]{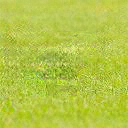}
        & \includegraphics[width=0.15\linewidth]{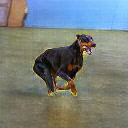}
        & \includegraphics[width=0.15\linewidth]{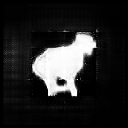}
        \\
    \rotatebox{90}{$\qquad$SEIGAN}
        & \includegraphics[width=0.15\linewidth]{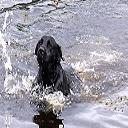}
        & \includegraphics[width=0.15\linewidth]{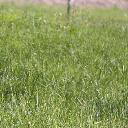}
        & \includegraphics[width=0.15\linewidth]{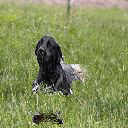}
        & \includegraphics[width=0.15\linewidth]{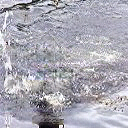}
        & \includegraphics[width=0.15\linewidth]{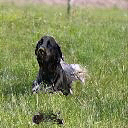}
        & \includegraphics[width=0.15\linewidth]{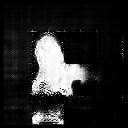}
        \\
    \rotatebox{90}{$\qquad$SEIGAN}
        & \includegraphics[width=0.15\linewidth]{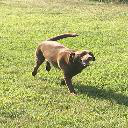}
        & \includegraphics[width=0.15\linewidth]{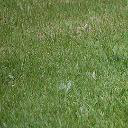}
        & \includegraphics[width=0.15\linewidth]{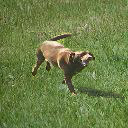}
        & \includegraphics[width=0.15\linewidth]{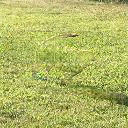}
        & \includegraphics[width=0.15\linewidth]{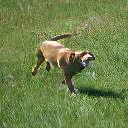}
        & \includegraphics[width=0.15\linewidth]{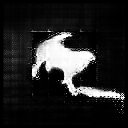}
        \\
\hline
    \rotatebox{90}{$\qquad$NoCycle}
        & \includegraphics[width=0.15\linewidth]{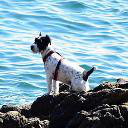}
        & \includegraphics[width=0.15\linewidth]{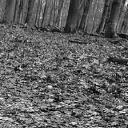}
        & \includegraphics[width=0.15\linewidth]{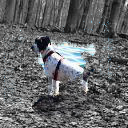}
        & % \includegraphics[width=0.15\linewidth]{images/examples_array/0b81b288_fake_B.png}
        & % \includegraphics[width=0.15\linewidth]{images/examples_array/0b81b288_fake_A.png}
        & \includegraphics[width=0.15\linewidth]{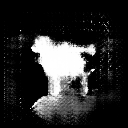}
        \\
\hline
\end{tabular}
\end{center}\vspace{-.5cm}
\caption{Samples of successful images and failures of our method and its variations.}
\label{fig:examples-array}
% \vspace{-1.5em}

% \begin{figure*}
    \begin{center}
    \newcolumntype{Y}{>{\centering\arraybackslash}X}
    %\resizebox{1\linewidth}{!}{%
    \begin{tabularx}{0.98\linewidth}{@{}YYYYYYYYYYYYYYYY@{}}
    O & T & N & S & O & T & N & S & O & T & N & S & O & T & N & S
    \end{tabularx}%}

        \includegraphics[width=1\linewidth]{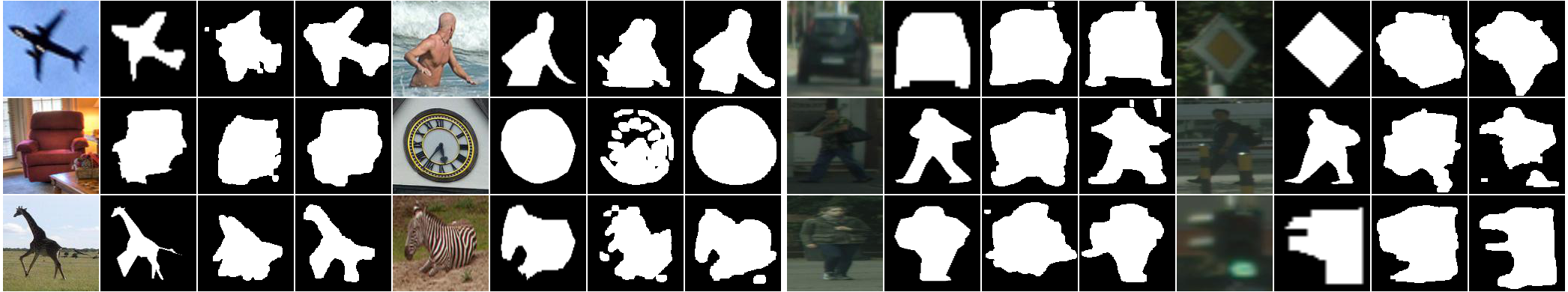}
    \end{center}\vspace{-.5cm}
    \caption{COCO and Cityscapes examples: original image (O), ground truth mask (T), masks predicted by NoCycle (N) and SEIGAN (S).}
    \label{fig:coco-samples}
% \end{figure*}
\end{figure*}

To partially address this, we introduce the so-called \emph{enhancement network} $\ge$ whose purpose is to generate a ``smoother'', more natural image $\hy=\langle \hO, \hB_y \rangle$ given the coarse image $\z=\m\odot\x + (1-\m)\odot \y=\langle O,B_y\rangle$ and random noise sampled from the standard normal distribution. We denote by $\ge(\z)$ the final improved image after all outputs of $\ge$ have been applied to $\z$ accordingly.
In our experiments with different $\ge$ architectures, we have not been able to achieve quite the level of background understanding to hide paws behind grass, but often $\ge$ can draw a shadow or a reflection under the dog (see Fig.~\ref{fig:architectures},~\ref{fig:examples-array}).

We train the swap network end-to-end with the following loss functions (shown by rounded rectangles on Fig.~\ref{fig:architectures}).

The \emph{object reconstruction loss} $\lro$ aims to ensure consistency and training stability. It is a weighted sum of perceptual loss and MAE between the source image $\x=\langle O, B_x \rangle$ and $\hhx = \ge(\gs(\hy)\odot\hy + (1-\gs(\hy))\odot \hx)$:
%
% $$\lro = \lambda_{\mathrm{rec_{percept}}} \left| \mathrm{VGG_2}(\x) - \mathrm{VGG_2}(\hhx) \right| + \left|\x - \hhx\right|,$$
\begin{align*}
  {\l_{\mathrm{obj}}^{\mathrm{perc}}} &= \left| \mathrm{VGG_2}(\x) - \mathrm{VGG_2}(\hhx) \right|,\quad
  {\l_{\mathrm{obj}}^{\mathrm{MAE}}} = \left| \x - \hhx \right|,\\
  \lro &= \lambda_6{\l_{\mathrm{obj}}^{\mathrm{perc}}} + 
  \lambda_7{\l_{\mathrm{obj}}^{\mathrm{MAE}}},
\end{align*}
where $\hy=\ge(z)$ and $\hx=\gi((1-\gs(\x))\odot\x)$, i.e., $\hhx$ is the result of applying the swap network twice.

The \emph{adversarial object loss} $\ldo$ aims to increase the plausibility of $\hy = \langle \hO, \hB_y \rangle$ and $z$. It is implemented with a dedicated discriminator network $\discro$. It also has the side effect of maximizing the area covered by segmentation mask $\m=\gs(\x)$. We apply this loss to images with objects: $\hy$ and $z$. Again, the discriminator has the same architecture as in CycleGAN except for the number of layers, where we have found that a deeper discriminator works better. We again use the MSE loss inspired by \mbox{LSGAN}~\cite{mao2016least}:
\begin{align*}
{\l_{\mathrm{coarse}}^{\mathrm{GAN}}} &= (1 - \discro(\z))^2,\quad {\l_{\mathrm{enh}}^{\mathrm{GAN}}} = (1 - \discro(\hy))^2,\\
\ldo &= \lambda_8{\l_{\mathrm{coarse}}^{\mathrm{GAN}}} + \lambda_9{\l_{\mathrm{enh}}^{\mathrm{GAN}}}.
\end{align*}

\def\lii{\l^{\mathrm{id}}}

Finally, apart from the loss functions defined above we have used the \emph{identity loss} $\lii$, an idea put forward in CycleGAN.
% We introduce two different instances of identity loss.
% \begin{itemize}
    % \item
    \emph{Object enhancement identity loss} $\lio$ brings the result of the enhancement network $\ge$ on real images closer to identity: it is the mean average distance between $\ge(\x)$ and $\x$ itself.
    % $\lio = \left|\ge(\x) - \x\right|;$
    % \item 
    \emph{Background identity loss} $\lib$ tries to ensure that the cut-and-inpaint procedure leaves intact images without objects: for an image $\y=\langle\emptyset,B_y\rangle$ we find a segmentation mask $\gs(\y)$, subtract it from $\y$ to get $(1-\gs(\y))\odot\y$, apply inpainting $\gi$ and then minimize the mean average distance between the original $\y$ and the result:
\begin{align*}
\lib &= \left|\gi((1-\gs(\y))\odot\y) - \y\right|,\\
\lio &= \left|\ge(\x) - \x\right|,\quad 
\lii = \lambda_{10}\lio + \lambda_{11}\lib.
\end{align*}
% \end{itemize}

\subsection{Total Loss Function, Remarks, and Network Architectures}

The  overall SEIGAN generator loss function is thus
% a linear combination of all loss functions defined above:
\begin{align*}
\l =& \lambda_1{\l_{\mathrm{inp}}^{\mathrm{GAN}}} +   \lambda_2{\l_{\mathrm{inp2}}^{\mathrm{GAN}}} +  \lambda_4{\l_{\mathrm{bg}}^{\mathrm{perc}}} + 
\lambda_5{\l_{\mathrm{bg}}^{\mathrm{MAE}}} + 
\lambda_6{\l_{\mathrm{obj}}^{\mathrm{perc}}} + \\
& \lambda_7{\l_{\mathrm{obj}}^{\mathrm{MAE}}} + 
\lambda_8{\l_{\mathrm{coarse}}^{\mathrm{GAN}}} + \lambda_9{\l_{\mathrm{enh}}^{\mathrm{GAN}}} + 
\lambda_{10}\lio + \lambda_{11}\lib,
% =& 
% \lambda_1{\l_{\mathrm{coarse}}^{\mathrm{gan}}} +
% \lambda_2{\l_{\mathrm{enh}}^{\mathrm{gan}}} +
% \lambda_3{\l_{\mathrm{inp}}^{\mathrm{gan}}} +
% \lambda_4{\l_{\mathrm{inp2}}^{\mathrm{gan}}} + \\
% &+\lambda_5\lrb +
% \lambda_6\lro +
% \lambda_7\lio +
% \lambda_8\lib,
\end{align*}
with coefficients $\lambda_1,\ldots,\lambda_{11}$ chosen empirically.

The discriminator losses are
\begin{align*}
\ldiscrb &=  (1 - \discrb(\y))^2 + \frac{1}{2} \discrb (\hx)^2 + \frac12\discrb(\hhy)^2,\\
\ldiscro &=  (1 - \discro(\x))^2 + \frac{1}{2} \discro (\hy)^2 + \frac12\discro(z)^2.
\end{align*}
To train SEIGAN, we train the generator and discriminators, alternating the minimization of $\l$ and $\ldiscrb+\ldiscro$.

To improve training stability, we used the experience replay technique originating from reinforcement learning~\cite{mnih2015human}, implemented in our case by maintaining a pool of fake images. After each mini-batch, we randomly update the pool and sample images from it for the next mini-batch.

Several interesting effects have emerged in our experiments. First, original images $\x=\langle O, B_x \rangle $ and $\y= \langle \emptyset, B_y \rangle $ might have different scale and aspect ratios before merging. Rescaling them to the same shape with bilinear interpolation would introduce significant differences in low-level textures that would be very easy to identify as fake for the discriminator, thus preventing GAN from convergence. The authors of~\cite{remez2018learning} faced the same problem and addressed it by a special procedure they use to create training samples: they took foreground and background patches only from the same image to ensure the same scale and aspect ratios, which reduces diversity and makes fewer images suitable for the training set. In our setup this problem is addressed by a separate enhancement network, so we have fewer limitations when looking for appropriate training data.

Another interesting effect is the low contrast in segmentation masks when inpainting is optimized against MAE or MSE reconstruction loss. A low-contrast mask (i.e., $\m$ with many values around $0.5$ rather than close to $0$ or $1$) allows information about the object to ``leak through'' and facilitate reconstruction. A similar effect has been noticed before by other researchers, and in the CycleGAN architecture it has even been used for steganography~\cite{chu2017cyclegan}. We first addressed this issue by converting the soft segmentation mask to a hard mask by thresholding, but later found that optimizing inpainting against the texture loss $\lrb$ is a more elegant solution that leads to better results than thresholding.

% {\bf I SUGGEST TO MAKE PICTURES OF THE ARCHITECTURES WHEN THEY ARE FINALIZED}

For the segmentation network $\gs$, we used the architecture from CycleGAN (Fig.~\ref{fig:arch}, right), which is an adaptation of the architecture from~\cite{DBLP:journals/corr/JohnsonAL16}.
% For better performance, we replaced \emph{ConvTranspose} layers with bilinear upsampling. 
After the final layer of the network, we used the logistic sigmoid as the activation function.
For the enhancement network $\ge$, we used the U-net architecture~\cite{DBLP:journals/corr/RonnebergerFB15} (Fig.~\ref{fig:arch}, left) since it can both work with high resolution images and make small changes in the source image. This is important for SEIGAN because we do not want to significantly change the image content in $\ge$ but rather just ``smooth'' the pasted image in a smarter way.

\begin{figure}
    \centering
    \setlength{\tabcolsep}{2pt}
    \begin{tabular}{cc}
        \includegraphics[width=0.49\linewidth]{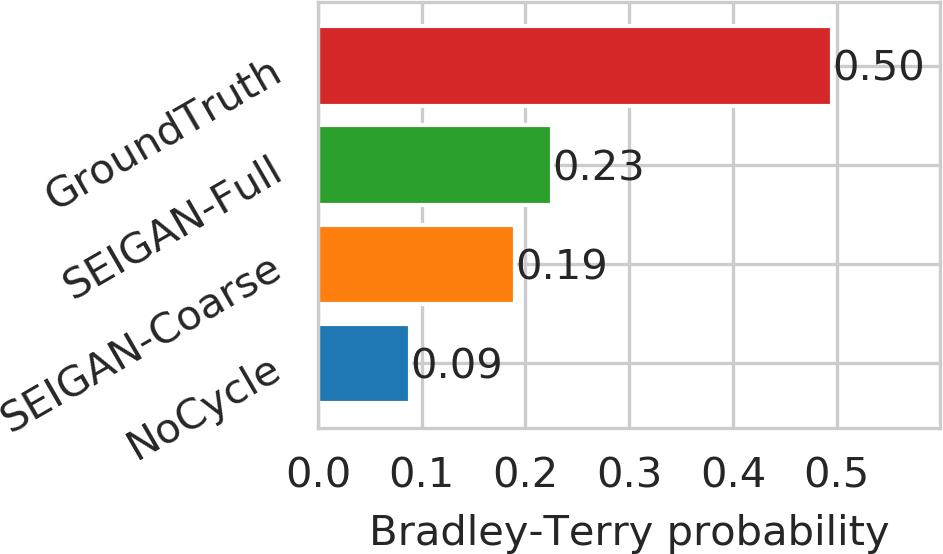}
        &
        \includegraphics[width=0.49\linewidth]{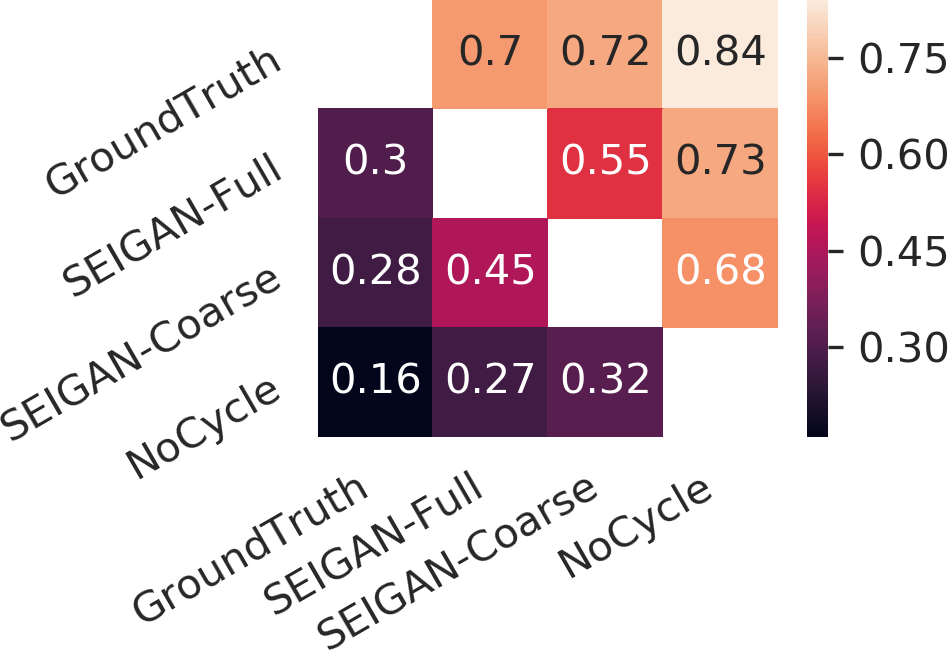}
    \end{tabular}
    
    \caption{Quality of generated images compared by human assessors; left: overall ratings estimated with the Bradley-Terry model; right: fractions of pairwise comparisons won (row vs. column).}
    \label{fig:ranking-plot}
    \vspace{-1.8em}
\end{figure}

\section{Experimental evaluation}\label{sec:eval}

In this section we present the results of our evaluation study, with two main experiments:
\begin{inparaenum}[(1)]
\item subjective assessment of generated images and
\item accuracy of generated segmentation masks measured against the ground truth.
\end{inparaenum}

\subsection{Subjective Assessment}\label{sec:crowd}

\textbf{Dataset.} We used the query ``dog'' to collect images from \emph{Flickr} (licensed under Creative Commons). Then we detected all objects (including dogs) and background-only regions without objects using Faster R-CNN (implementation from~\cite{jjfaster2rcnn}, pretrained on COCO, ResNet-101 backbone). Then we constructed two datasets: one from regions with dogs $X= \left\{ \langle O_x, B_x \rangle\right\}_{\x\in X}$ and another from background-only regions $Y = \left\{ \langle \emptyset, B_y \rangle\right\}_{\y\in Y}$. After data collection, we filtered the regions according to the following rules:
\begin{inparaenum}[(i)]
    \item after rescaling, the object size is equal to $64\times 64$ and size of the final crop is equal to $128\times 128$;
    \item the object is located at the center of the crop;
    \item there are no other objects that intersect with a given crop;
    \item the source size of the object on a crop exceeds $60$px (by the smaller side) and does not exceed $40$\% of the entire source image (by the longer side).
\end{inparaenum}
We used this dataset for the main part of our experiments with architectures and loss functions, as well as for subjective image quality assessment.

\textbf{Crowdsourcing Evaluation Procedure.} This experiment aims to evaluate the impact of different architectures and loss functions on the quality of the resulting image $\hx$. We prepared four sets of 100 images each: 
% \vspace{-0.8em}
% \begin{itemize} 
\begin{inparaenum}[(1)]
    \item \emph{GroundTruth}: real images with dogs and some background; %\vspace{-0.8em}
    \item \emph{NoCycle}: coarse images $\z$ obtained after training our model with only $ \ldo $ enabled (similar to the pipeline in~\cite{remez2018learning}); %\vspace{-0.8em}
    \item \emph{SEIGAN-Coarse}: coarse images $ \z $ obtained after training our model with all loss functions enabled ($\ge$ not applied); %\vspace{-0.8em}
    \item \emph{SEIGAN-Full}: images $\hx$ obtained after training our model with all loss functions enabled ($\ge$ applied). 
    % \vspace{-0.8em}
\end{inparaenum} 
% \end{itemize}
%
Images for each set were selected according to their corresponding discriminator score (except \emph{GroundTruth}, which is a uniform subsample of $X$).

Then we collected $9000$ pairwise comparisons of images randomly sampled from different sets via the crowdsourcing platform \emph{Yandex.Toloka}\footnote{\url{https://toloka.yandex.com}}. We compared a total of $1200$ images ($300$ per set), each pair was assessed by $5$ participants, with $458$ human assessors involved in total.

\textbf{Results.} 
We estimated the ratings of models based on pairwise comparisons using the Bradley-Terry model~\cite{bradley1952rank}; the ratings and more detailed pairwise results are shown on Figure~\ref{fig:ranking-plot}. Fig.~\ref{fig:ranking-plot} clearly shows that \textit{SEIGAN} produces significantly more realistic images than \textit{NoCycle}; \textit{SEIGAN-Full} was even able to win $30$\% of comparisons against \textit{GroundTruth}.
Interestingly, the refinement network in \textit{SEIGAN-Full} rarely improves the image in terms of perceptual quality over simple mask blending. However, it facilitates training and allows \textit{SEIGAN-Coarse} to win against \textit{NoCycle} due to a much more accurate segmentation mask. This is also confirmed by our segmentation experiments shown below. Thus, the large gap between \textit{GroundTruth} and \textit{SEIGAN-Full} can be explained by the fact that we did not specifically select backgrounds $\y$ to match the dogs $\x$, and in some cases it was impossible to paste that particular dog onto that particular background and make the image look natural (the second example in Fig.  ~\ref{fig:examples-array}).

\begin{table}
\begin{center}
\begin{tabular}{c|c|c}
\hline
    Algorithm & COCO & Cityscapes \\
\hline
    Unsupervised NoCycle  & 0.723 & 0.754 \\
    Unsupervised SEIGAN      & \textbf{0.762} & \textbf{0.802} \\
\hline
    Supervised  & 0.83  & 0.867 \\
\hline
\end{tabular}
\end{center}
\caption{Intersection-over-Union scores of object segmentation masks against the ground truth (in original resolution); predicted masks were binarized with confidence threshold $0.5$.}
\label{table:segmentation-results}
\vspace{-2.5em}
\end{table}

\subsection{Weakly Supervised Object Segmentation}\label{sec:segm}

We have experimented with weakly supervised segmentation in order to numerically estimate the quality of object segmentation masks produced by our method on heterogeneous data. We have conducted the experiments on COCO~\cite{lin2014microsoft} and Cityscapes~\cite{Cordts2016Cityscapes} datasets. All experiments were performed on the full set of classes, and the trained models were not aware of target class labels.

The data preparation procedure is similar to the one defined in Section~\ref{sec:crowd}, with the following differences. The source size of the object on a crop has to be larger than $20$px (by the smaller side) and no larger than that $40$\% of the source image (by the larger side). The aspect ratio of an object (smaller side divided by the larger side) has to exceed 0.4. With scaling, we reduced all object sizes to $100\times 100$.

Segmentation masks are compared against ground truth masks in the original resolution. For this purpose, the predicted mask of size $128 \times 128$ is resized to the original shape with bilinear interpolation, preserving the original aspect ratio. Note that for cut-and-paste algorithms it may be useful to include in the segmentation mask not only the object itself but some parts of the surrounding area, e.g., the object's shadow. This may lead to a slightly lower final IoU score but still improve the overall quality of the image.

We evaluated SEIGAN against two baselines: \emph{Supervised}, a segmentation network of the same architecture as $\gs$ but trained with ground truth segmentation masks, and \textit{NoCycle} (same as in Section~\ref{sec:crowd}). To train the \emph{Supervised} baseline, we used the labeled ``train'' subsets of the corresponding datasets; all metrics were computed on ``test'' subsets. We counted images with objects ($X$) in the datasets after preprocessing; there were at least twice as many images without objects ($Y$) for each subset.
%
% \textbf{Cityscapes.} 
To train \textit{SEIGAN} and \textit{NoCycle} on {Cityscapes}, we used all data from leftImg8bit\_sequence for training ($77963$ images) and (test, val) from annotated part for validation ($1518$ images). The supervised model was trained on the ''train'' subset of annotated data ($8963$ images).
% \textbf{COCO.} 
On the COCO dataset, all models were trained with the original train/validation split. The train set contains $31176$ images with \textit{person} ($4968$ images), \textit{clock} ($2766$), and \textit{traffic light} ($1782$) being the most common classes. The validation set contains $1329$ images. 

Fig.~\ref{fig:coco-samples} shows sample segmentation masks for both Cityscapes and COCO, and
% Some examples of generated masks may be seen in Fig. \ref{fig:coco-samples}.
% \textbf{Results.} 
evaluation results are presented in Table~\ref{table:segmentation-results}. It clearly shows that \textit{SEIGAN} achieves competitive intersection-over-union scores compared even to the supervised model. Both on COCO and Cityscapes, it significantly outperforms \textit{NoCycle}.
% , generating more accurate segmentation masks.

%------------------------------------------------------------------------
\section{Conclusion}\label{sec:concl}

Compositional image generation is an approach to creating new images by combining parts of existing images, using machine learning models to perform the required cut-and-paste operations in a way consistent with image semantics. In this work, we have proposed a novel approach to compositional image generation based on SEIGAN, a new end-to-end model for segmentation, enhancement, and inpainting. Apart from compositional image generation, our model produces high quality segmentation masks as a side result of the cut-and-paste pipeline, as evidenced by our experiments both for segmentation and for image quality evaluated by human assessors. However, both SEIGAN and other state of the art models still work only with a single object against an empty background. Therefore, as open problems for future work we highlight learning to insert objects on backgrounds that have other objects, choosing a place to paste the object, constructing depth maps for this purpose and to improve scaling, and training the model without separately produced bounding boxes.

{\small
\bibliographystyle{ieee}
\bibliography{references}
}

\end{document}